\documentclass[11pt]{article}

\usepackage[preprint]{acl}

\usepackage{times}
\usepackage{latexsym}
\usepackage{booktabs}

\usepackage{subcaption}   
\usepackage[utf8]{inputenc}
\usepackage[T1]{fontenc}
\usepackage{textalpha}
\usepackage[T1]{fontenc}

\usepackage[utf8]{inputenc}

\usepackage{microtype}

\usepackage{inconsolata}

\usepackage{graphicx}
\usepackage{booktabs}
\usepackage{makecell}
\usepackage{tabularx}
\usepackage{float}
\DeclareFontFamilySubstitution{LGR}{ptm}{artemisia}
\DeclareFontFamilySubstitution{LGR}{zi4}{cmtt}

\title{More Capable, Less Faithful: A Multilingual Analysis of Mathematical (Un)Solvability Detection in LLMs}

\author{
\begin{tabular}{ccc}
Maria-Eleni Zoumpoulidi\thanks{Equal contribution. Author order determined by coin flip.} &
Nikolaos Xiros\footnotemark[1] &
Georgios Paraskevopoulos
\end{tabular}
\\[0.6em]
{\normalfont Institute for Language and Speech Processing, Athena Research Center, Greece}\\
{\normalfont
\texttt{mzoumpoulidi@gmail.com} \quad
\texttt{\{n.xiros,g.paraskevopoulos\}@athenarc.gr}
}
}

\begin{document}
\maketitle

\begin{abstract}
Solvability detection is one of the most challenging aspects of mathematical reasoning for Large Language Models (LLMs). While prior work has studied this capability extensively, these analyses have been limited to English. Consequently, it remains unclear whether multilingual failures arise from differences in internal Solvability Belief or from language-dependent failures to express it. To address this gap, we introduce the first multilingual benchmark of paired solvable and unsolvable mathematical problems, extending ReliableMath to French and Greek. Using this, we train multilingual probes predicting Solvability Belief and analyze the solvability detection capabilities of state-of-the-art LLMs behaviorally, representationally, and in terms of faithfulness. We find that Solvability Belief is encoded as a largely universal, language-agnostic feature, and that higher-resource languages such as English, despite achieving stronger mathematical reasoning performance, exhibit lower solvability-detection faithfulness.
\end{abstract}

\section{Introduction}
Studying the mathematical capabilities of Large Language Models (LLMs) across multiple languages is an important and growing research direction, aimed at improving performance across languages.

Research on multilingual mathematical reasoning has focused primarily on accuracy over solvable problems. Existing work spans benchmarks (e.g., \citet{shi2023language}, \citet{xu2026mgsmprosimplestrategyrobust}), training approaches \citep{chen2024breakinglanguagebarriersmultilingual}, and prompting strategies such as \citet{shi2023language}, and \citet{barua2026longchainofthoughtreasoninglanguages}. Representation-level studies investigate the encoding of problem difficulty \citet{civelli-etal-2026-shared}. However, multilingual solvability detection remains underexplored.

The picture is different for solvability detection in English. Prior work covers benchmarks and in-context learning (e.g., \citet{xue2025reliablemathbenchmarkreliablemathematical}, \citet{sun-etal-2024-benchmarking}) as well as investigations of models' internal representations. More specifically, \citet{xiros2026knowledgeknowsverbalizationtells} and 
\citet{sanyal2025confidencecompetence} 
disentangle Solvability Belief and verbalization in LLMs, showing that they correspond to distinct, geometrically decoupled directions. Furthermore, \citet{liu2026answeringunanswerableerrknowingly} show that LLMs often internally recognize unsolvable problems despite failing to abstain.

However, none of these works investigates whether models' internal Solvability Belief transfers faithfully across languages or whether this belief is verbalized consistently in multilingual settings. Consequently, it remains unknown whether multilingual mathematical reasoning failures stem from differences in internal knowledge or from language-dependent failures to express it.
To address this gap, we construct a multilingual version of ReliableMath \citep{xue2025reliablemathbenchmarkreliablemathematical} by translating both its solvable problems and their unsolvable counterparts into French and Greek. Using this benchmark, we investigate how language affects both models' internal representations of solvability and the faithfulness with which these representations are reflected in their generated responses.

We aim to answer the following research questions: \textbf{RQ1:} How does LLM performance in mathematical reasoning and unsolvability detection vary across languages? \textbf{RQ2:} How does language affect the faithfulness between internal solvability representations and textual solvability judgments?

Our main contributions are:
\begin{itemize}
\item We introduce the first multilingual benchmark for paired solvable-unsolvable mathematical problems, extending ReliableMath to French and Greek. Our code and data will be available under the Apache 2.0 license. \footnote{Data: \href{https://anonymous.4open.science/r/MoreCapableLessFaithful-0F5E/}{repository}.}
\item We provide a behavioral analysis of multilingual mathematical reasoning and unsolvability detection across state-of-the-art LLMs.
\item We present the first multilingual representational analysis of Solvability Belief and its faithfulness to textual outputs, showing that stronger languages tend to exhibit lower solvability-detection faithfulness.
\end{itemize}

\section{Methodology}
Our methodology comprises four stages: benchmark translation, chain-of-thought (CoT) generation and hidden-state extraction, probing of solvability representations and LLM-based annotation of the output text's solvability verdict. Each stage is described in detail below.

\paragraph{Benchmark Translation:}
We translate the ReliableMath benchmark \citep{xue2025reliablemathbenchmarkreliablemathematical} (the prompt used can be found in Appendix~\ref{sec:app_translation_prompt}), including both solvable problems and their unsolvable counterparts, into French and Greek using Claude Sonnet 5 \citep{anthropic2026claudesonnet5}. We perform human validation to ensure that mathematical content, numerical values, and solvability type are preserved (Appendix ~\ref{sec:appendix_human_annotation_translate}).

\paragraph{CoT generation and hidden-state extraction:}
For each language, we generate Chain-of-Thought (CoT) responses using language-specific prompt templates (Appendix~\ref{sec:app_cot_prompts}). We consider two prompting settings: \textit{standard}, which instructs the model to solve the problem step by step, and \textit{aware}, which additionally allows it to explicitly state that a problem is unsolvable. We use the hidden states extracted from the layer yielding the highest probing performance (Appendix~\ref{sec:app_Methodology_details}), with each sequence represented by 20 uniformly sampled token vectors.

\paragraph{Probing:} We probe Solvability Belief (SB), using the ground-truth solvability label of the problem as a proxy for the model's internal belief. Hidden-state representations from the standard and solvability-aware prompting settings are pooled to train a single L1-regularized logistic regression probe for predicting SB.

\paragraph{Textual Solvability Verdict Annotation:} Determining the textual solvability verdict is non-trivial, as reasoning traces for unsolvable problems often contain shifts in the model's assessment throughout the generation process. Rather than relying on stricter alternatives, such as matching predefined phrases, we adopt an LLM-as-a-judge approach using Llama-3.3-70B-Instruct \citep{grattafiori2024llama3herdmodels}. The judge infers the dominant solvability verdict conveyed by the reasoning trace, distinguishing between solution-oriented reasoning and recognition of unsolvability (the corresponding prompt can be found in Appendix~\ref{sec:app_judge_prompt}). The judge was validated through human annotation on a subset of 100 samples, achieving a high agreement score (see Appendix~\ref{sec:appendix_human_annotation}).

\section{Experiments}
\paragraph{Models:} For our experiments, we query Qwen3-30B-A3B-Instruct-2507, Qwen3-4B-Instruct-2507, \citep{yang2025qwen3technicalreport}, Llama-3.1-8B-Instruct \citep{grattafiori2024llama3herdmodels}, Gemma4-31B-IT \citep{gemma4_modelcard_2026}. Additionally, we include two language-specialized variants of Llama-3.1-8B-Instruct: Llama-Krikri-8B \citep{roussis-etal-2025-krikri}, adapted to Greek, and French-Alpaca \citep{pacifico2024frenchalpaca}, adapted to French. For these two models, we evaluate only on English and their respective specialization language.


\paragraph{Datasets:} We use our translated version of ReliableMath, which comprises 313 solvable problems drawn from MATH \citep{hendrycksmath2021} (high-school level), MinervaMath \citep{mathai_minervamath} (competitive college level), and AIME24/AMC \citep{aimo2024a, aimo2024b} (Olympiad level), together with 1,102 unsolvable variants obtained by removing or contradicting information essential to the solution.

\section{Results}
\label{sec:results}

\subsection{Models Achieve Strongest Mathematical Reasoning On Their Dominant Language}
\label{sec:baseline-capability}

We first benchmark how accurately each model solves problems on the solvable split of our dataset, per language, under a standard prompt, in order to measure raw task capability across languages.

Table~\ref{tab:standard-prompt-accuracy} shows a clear capability hierarchy that is largely independent of language: Gemma-4-31B-it and both Qwen3 variants substantially outperform the Llama-based models across all three languages. In nearly every model, performance is best in English and degrades as we move to French and further to Greek. This pattern is
consistent with our expectation that models reason more effectively in the language dominating their pretraining data.

\begin{table}[h]
\centering
\small
\begin{tabular}{lccc}
\toprule
\textbf{Model} & \textbf{English} & \textbf{French} & \textbf{Greek} \\
\midrule
Gemma-4-31B-it              & \textbf{61.3} & 55.6 & 56.5 \\
Qwen3-4B-Instruct           &  52.1 &  \textbf{53.3} & 43.5 \\
Qwen3-30B-Instruct          & \textbf{54.3} & 53.7 & 49.8 \\
Llama-3.1-8B-Instruct       & \textbf{23.3} & 15.0 & 14.4 \\
French-Alpaca-Llama3-8B     & \textbf{7.7}  & 7.4  & --  \\
Llama-Krikri-8B-Instruct    & \textbf{17.6} & -- & 14.7 \\
\bottomrule
\end{tabular}
\caption{Accuracy (\%) on the solvable split under the standard prompt, by model and language.}
\label{tab:standard-prompt-accuracy}
\end{table}

\subsection{Internal Solvability Belief Is Strongest In The Dominant Language, But Highly Universal}
\label{subsec:knowledge_probe}

Moving from raw accuracy on solving problems, we next assess the validation AUC of the Solvability Belief (SB) probe, which measures how sharply a model's hidden states encode the solvability of a problem internally. We evaluate this across each of the three languages individually, using both per-language probes and a pooled \emph{universal} probe trained jointly on all available languages.

\begin{figure}[h]
\centering
\includegraphics[width=\columnwidth]{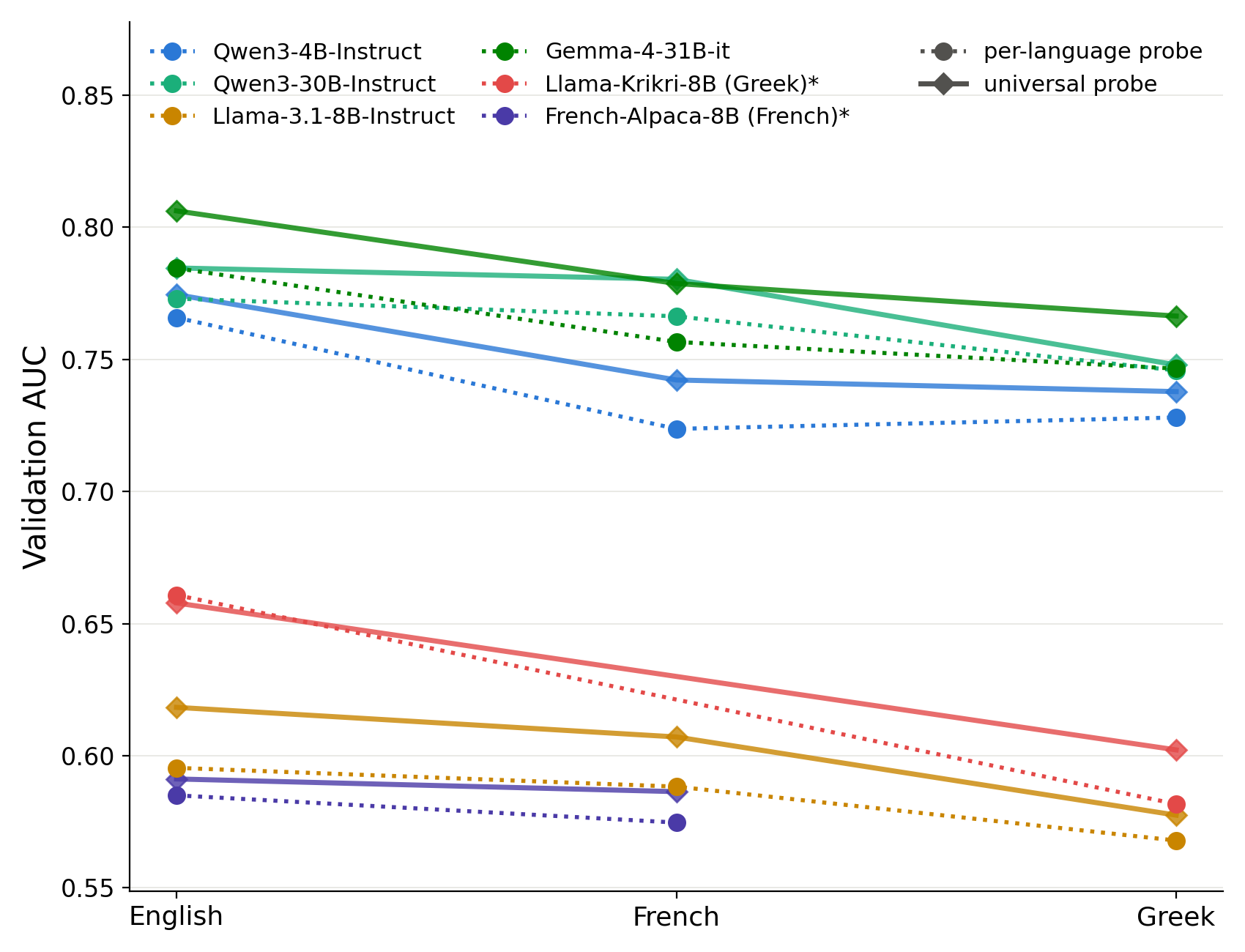}
\caption{Validation AUC by language for the Solvability Belief probe. Probe AUC degrades monotonically from English to Greek for every model. *=language-specialized.}
\label{fig:probe-auc-combined}
\end{figure}

\paragraph{Solvability Belief is less encoded in underrepresented languages}
Figure~\ref{fig:probe-auc-combined} plots validation AUC by language for both probe types. Across all six models, AUC point estimates are highest in English and decline as we move to French and further to Greek, for both the per-language and universal probes. Notably, the drop also persists for the two language-specialized models (Llama-Krikri-8B and French-Alpaca-8B), suggesting that the dominant pretraining language outweighs the language-specific fine-tuning. All Llama-based models, having weaker baseline reasoning capability as shown before, sit lowest throughout. The sharper encoding of Solvability Belief in English is consistent with the stronger reasoning performance observed for English in Section~\ref{sec:baseline-capability}. Significance testing details are provided in Appendix~\ref{app:sb_significance}.

\paragraph{Solvability Belief Probe is highly universal}
As shown in Figure~\ref{fig:probe-auc-combined}, pooling all available languages into a single \emph{universal} probe matches or exceeds the per-language probe performance for every model. A bootstrap check confirms that in most cases this improvement is statistically significant (Appendix~\ref{app:sb_significance}). This indicates the Solvability Belief signal is largely encoded in a shared, only mildly language-specific subspace. Thus, we adopt the \emph{universal} probe for our analysis.



\begin{figure*}[h]
    \centering
    \begin{minipage}{0.49\textwidth}
        \centering
        \includegraphics[width=\linewidth]{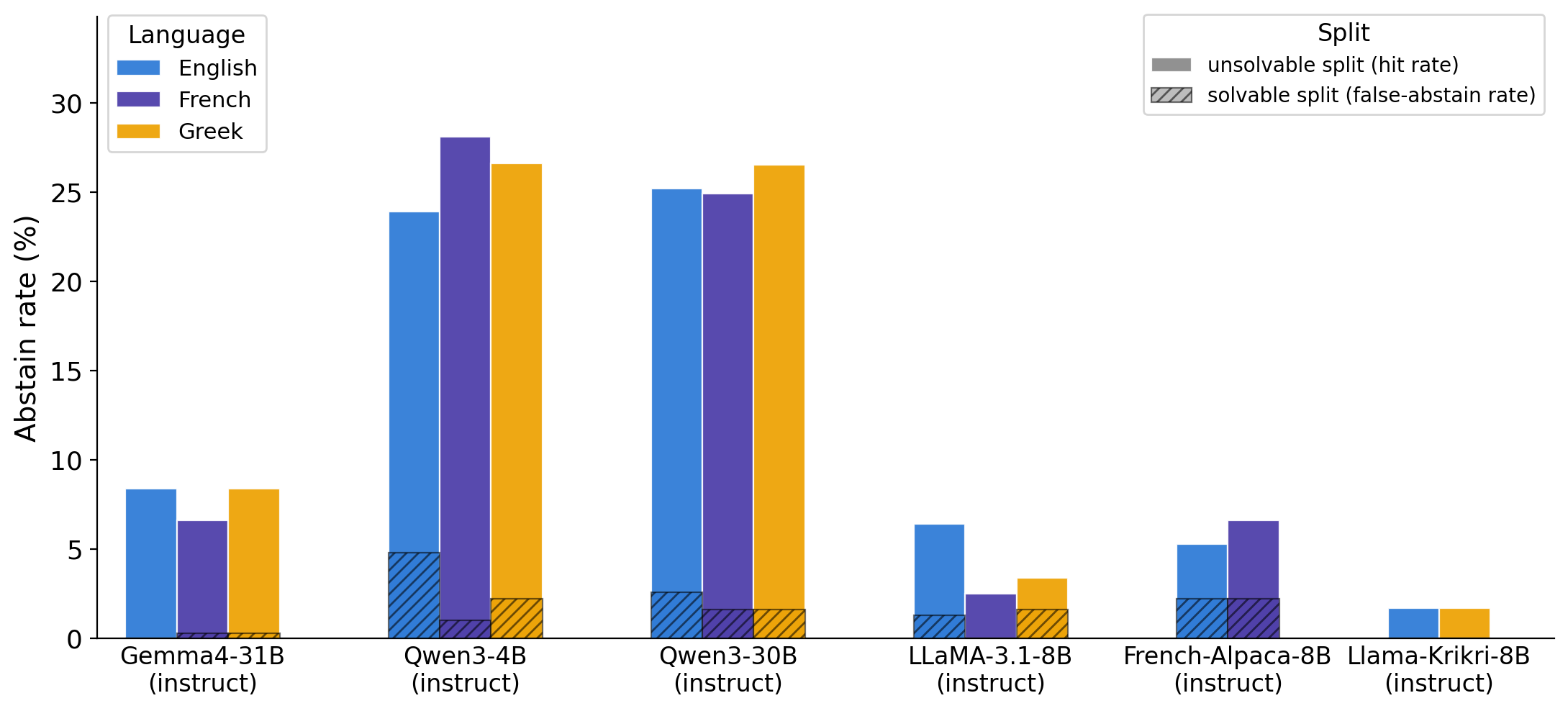}
        \subcaption{Standard prompt}
        \label{fig:abstain_rates_standard}
    \end{minipage}
    \hfill
    \begin{minipage}{0.49\textwidth}
        \centering
        \includegraphics[width=\linewidth]{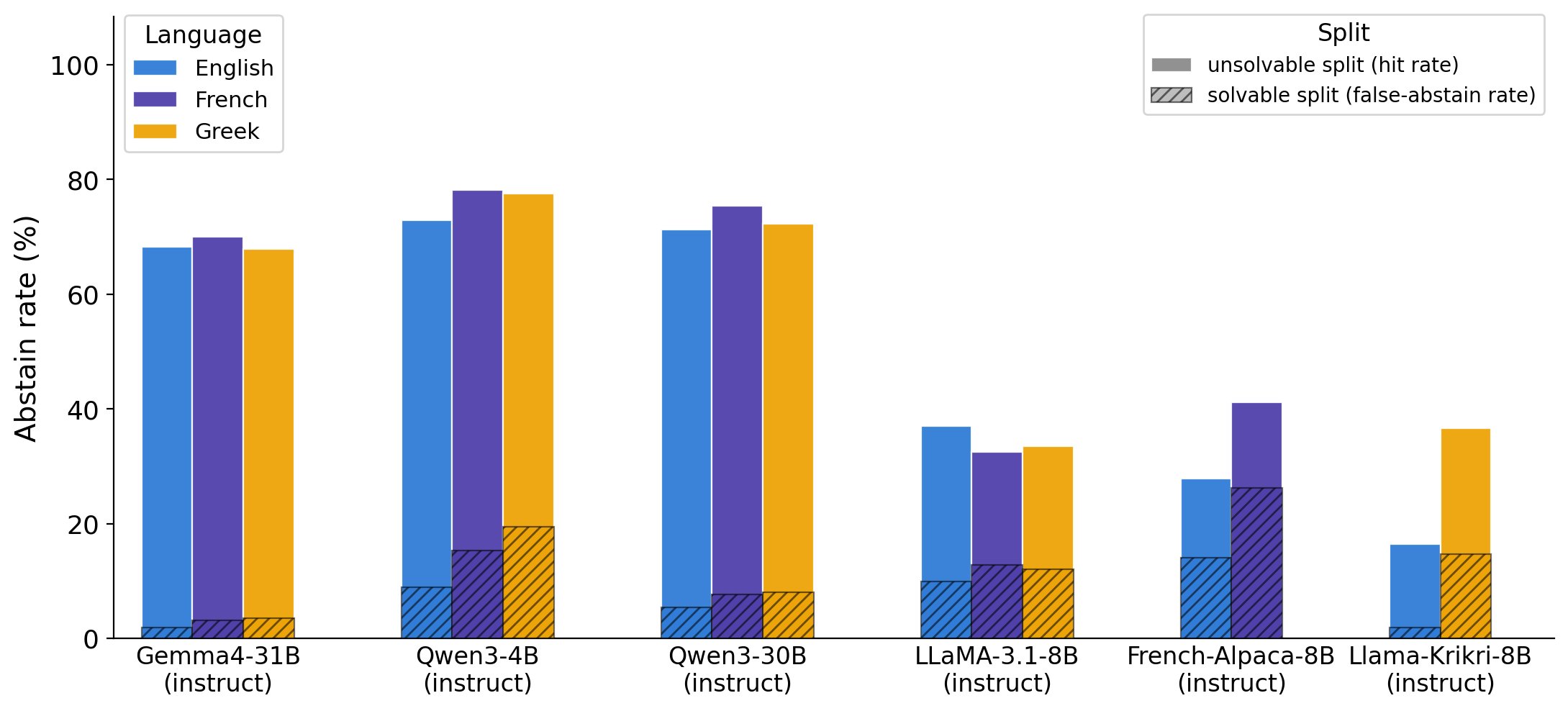}
        \subcaption{Aware prompt}
        \label{fig:abstain_rates_language_aware}
    \end{minipage}
    \caption{Abstention rates (\%) across English, French, and Greek for various language models. The segments distinguish between the hit rate (true abstention) on the unsolvable split and the false-abstentions rate on the solvable split for the standard prompt (left) and the aware prompt (right).}
    \label{fig:abstain_rates}
\end{figure*}

\subsection{Language Adaptation Raises Abstention}
\label{subsec:abstain}

Having studied how models encode solvability internally, we now measure the behavioral abstention profiles across models and languages. Figure~\ref{fig:abstain_rates} reports the percentage of true abstentions on the unsolvable split alongside false-abstentions (false refusals) on the solvable split, under both prompt settings.

Under the standard prompt (Figure~\ref{fig:abstain_rates_standard}), abstention is low for most models: only the two Qwen3 variants abstain at substantial rates ($\sim$25\%), with Llama-Krikri-8B barely abstaining at all. Notably, under the standard prompt, the two Qwen3 variants perform \emph{better} in the underrepresented languages than in English: both Qwen3-4B and Qwen3-30B show a higher hit rate on the unsolvable split and a lower false-abstentions rate on the solvable split in Greek than in English. 

The aware prompt raises abstention sharply for every model (Figure~\ref{fig:abstain_rates_language_aware}). The natively multilingual models (Gemma-4-31B, Qwen3 variants) flag unsolvable problems at a comparable rate across all three languages, but their false-abstentions rate rises as we move from English to French and Greek. Llama-3.1-8B, with virtually no Greek pretraining, abstains markedly better in English than in French or Greek. Its language-adapted variants reverse this: French-Alpaca-8B abstains most in French and Llama-Krikri-8B in Greek. Crucially, in both models, this gain is not free: the surge of true abstentions in the target language is accompanied by a spike in false-abstentions concentrated in that same language. Thus, under aware prompting, language adaptation appears to lower the abstention threshold globally for the target language, increasing both true hits and false refusals.

\subsection{Models Are Least Faithful In Their Dominant Language}

Having found that models reason more capably in their dominant language (Section~\ref{sec:baseline-capability}) and encode Solvability Belief more sharply in that same language (Section~\ref{subsec:knowledge_probe}), yet do not verbalize that belief any more reliably (Section~\ref{subsec:abstain}), we are motivated to directly measure faithfulness across languages. For each sample, we compare the universal Solvability Belief probe's prediction against the textual verdict assigned by the LLM judge, and define their agreement rate as the \textbf{faithfulness rate}.

Table~\ref{tab:faith-pooled}, which reports the overall faithfulness rate across all prompt styles and solvability splits, shows a striking reversal of the pattern observed for capability and internal belief: for the natively multilingual models (Gemma and Qwen), faithfulness is \emph{higher} in the underrepresented languages than in English, whereas the least multilingual Llama-3.1-8B shows the opposite ordering, with faithfulness highest in English and lowest in Greek. The two language-specialized models follow the same underrepresented-language pattern: both Krikri-8B and Alpaca-8B are more faithful in their respective specialization language (Greek and French) than in English.

\begin{table}[h]
\centering
\small
\setlength{\tabcolsep}{4pt}
\begin{tabular}{lccc}
\toprule
Model & En & Fr & Gr \\
\midrule
Gemma-4-31B   & 0.612 & 0.615 & \textbf{0.626} \\
Qwen3-4B      & 0.691 & \textbf{0.728} & \textbf{0.728} \\
Qwen3-30B     & 0.712 & \textbf{0.730} & 0.723 \\
Llama-3.1-8B  & \textbf{0.544} & 0.529 & 0.498 \\
Alpaca-8B    & 0.473 & \textbf{0.498} & --    \\
Krikri-8B    & 0.500 & --    & \textbf{0.510} \\
\bottomrule
\end{tabular}
\caption{Faithfulness rate, pooled across prompt styles and solvability types.}
\label{tab:faith-pooled}
\end{table}

To better understand this effect, we restrict our faithfulness analysis to the unsolvable split and compare standard$\rightarrow$aware prompt pairs (Table~\ref{tab:faith-unsolv}). As shown, the aware prompt raises faithfulness for every model, but far more so in underrepresented languages: the gain is smallest in English and largest in French or Greek across all six models. This suggests that explicitly prompting models to consider unsolvability closes the gap between internal belief and verbalized output more effectively in lower-resource languages than in English. Significance testing details are provided in Appendix~\ref{app:significance}.

\begin{table}[h]
\centering
\small
\setlength{\tabcolsep}{3pt}
\begin{tabular}{lccc}
\toprule
Model & En & Fr & Gr \\
\midrule
Gemma-4-31B   & 0.271$\to$0.762 & 0.253$\to$\textbf{0.791} & 0.287$\to$0.787 \\
Qwen3-4B      & 0.484$\to$0.773 & 0.532$\to$0.815 & 0.535$\to$\textbf{0.832} \\
Qwen3-30B     & 0.509$\to$0.796 & 0.524$\to$0.819 & 0.506$\to$\textbf{0.829} \\
Llama-3.1-8B  & 0.400$\to$0.510 & 0.374$\to$0.514 & 0.305$\to$\textbf{0.523} \\
Alpaca-8B    & 0.330$\to$0.434 & 0.308$\to$\textbf{0.558} & --              \\
Krikri-8B    & 0.375$\to$0.413 & --              & 0.273$\to$\textbf{0.556} \\
\bottomrule
\end{tabular}
\caption{Faithfulness rate on unsolvable problems, standard$\rightarrow$aware prompt.}
\label{tab:faith-unsolv}
\end{table}

\section{Conclusions}
We introduced the first multilingual version of ReliableMath, extended to French and Greek, and used it to study solvability detection both behaviorally and through models' internal representations. We find a consistent capability hierarchy, with models performing best in English and worst in Greek, but the opposite pattern for faithfulness: multilingual models are more faithful to their internal Solvability Belief in underrepresented languages than in English. Our findings suggest that a model's dominant language yields stronger raw capability but not more trustworthy self-reports.

\section*{Limitations}
We acknowledge that, while our analysis provides informative insights, it would benefit from broader language coverage. Future work will extend our analysis to additional high-resource languages (e.g., Chinese) and typologically diverse, lower-resource languages. We also plan to leverage these insights to develop methods that better align textual judgments with models' internal Solvability Beliefs, improving the reliability of multilingual solvability detection.

\section*{Ethical Considerations}

This work is purely analytical: we evaluate and probe existing publicly available models and do not release any new model, nor propose any method intended for deployment. Our benchmark is a direct translation of ReliableMath \citep{xue2025reliablemathbenchmarkreliablemathematical} into French and Greek; the underlying problems are mathematical in nature and contain no personal, sensitive, or offensive content, and the translation preserves the original items without adding, removing, or reweighting any of them. 

\bibliography{custom}

\appendix

\section{Translation Prompt}
\label{sec:app_translation_prompt}
The prompt used for the dataset translation can be found in Table ~\ref{tab:translation-prompt}.

\section{CoT Prompts}
\label{sec:app_cot_prompts}
We present the prompts used for chain-of-thought (CoT) generation across all three languages, for both the standard and unsolvable-aware prompting settings. The standard prompts can be found in Table~\ref{tab:standard-prompts}, while the unsolvable-aware prompts in Table~\ref{tab:aware-prompts}.

\section{Judge Prompt}
\label{sec:app_judge_prompt}
The prompt used by the LLM judge to annotate models' verbal responses is shown in Table~\ref{tab:judge-prompt}.

\section{Layer selection}
\label{sec:app_Methodology_details}

In Table~\ref{tab:layers}, we report the selected layer for each model, which corresponds to the one achieving the highest probe AUC. For all models, these correspond to the middle or middle-to-late layers.

\begin{table}[H]
\centering
\begin{tabular}{lc}
\toprule
\textbf{Model} & \textbf{Layer} \\
\midrule
Gemma-4-31B-it                 & 36 \\
Qwen3-4B-Instruct                      & 18 \\
Qwen3-30B-Instruct                     & 36 \\
Llama-3.1-8B-Instruct/ KriKri / Alpaca                 & 16 \\
\bottomrule
\end{tabular}
\caption{Selected layer used for our experiments. For each model, we use the layer with the highest probing performance.}
\label{tab:layers}
\end{table}

\section{Significance testing details}





\subsection{Significance of Solvability Belief Results}
\label{app:sb_significance}

We compute problem-level bootstrap confidence intervals (1000 resamples, 95\% level) for the two claims in Section~\ref{subsec:knowledge_probe}. Since English, French, and Greek data are parallel translations of the same problems, we use a paired bootstrap, resampling the same problems across languages for direct comparison. All effect sizes below are AUC differences reported $\times 100$.

\paragraph{English vs.\ other languages}
The English advantage over Greek is significant for Qwen3-4B ($+3.8$, CI $[1.3, 6.5]$), Qwen3-30B ($+2.8$, CI $[0.5, 5.2]$), Gemma-4-31B ($+3.8$, CI $[1.8, 6.0]$), and Llama-Krikri-8B ($+7.9$, CI $[4.8, 10.7]$). Notably, Krikri has the largest gap we observe, despite being Greek-adapted. The advantage over French is significant only for Qwen3-4B ($+4.2$, CI $[2.1, 6.5]$) and Gemma-4-31B ($+2.8$, CI $[1.0, 4.7]$). For Llama-3.1-8B and French-Alpaca-8B, point estimates favor English throughout ($+0.8$ to $+2.8$) but all CIs cross zero. French vs.\ Greek is never significant for any model (largest gap $+2.1$ for Qwen3-30B, CI $[-0.1, 4.1]$).

\paragraph{Universal vs.\ per-language probes}
The universal probe is never significantly worse than the per-language probe across any of the 16 model-language combinations tested. It is significantly better for Gemma-4-31B in all three languages ($+2.0$ to $+2.2$), for Qwen3-4B, Qwen3-30B, and Llama-3.1-8B in both English and French ($+0.9$ to $+2.3$), and for Llama-Krikri-8B in Greek ($+2.1$, CI $[0.8, 3.5]$). The remaining comparisons do not reach significance, but every point estimate stays positive (largest such: $+1.2$ for French-Alpaca-8B in French, CI $[-0.8, 3.5]$). Pooling is thus at worst neutral and at best beneficial throughout.

\subsection{Significance testing for faithfulness claims}
\label{app:significance}

We test both faithfulness comparisons with a two-proportion $z$-test on
agreement counts. For the pooled comparison, Qwen3-4B's English-worse
effect is significant ($p<0.001$), as is Llama-3.1-8B's reversed,
English-better effect ($p<0.01$); Qwen3-30B, Gemma-4-31B, Krikri-8B, and
Alpaca-8B trend in the same English-worse direction but do not reach
significance ($p$ between 0.07 and 0.49). For the aware-prompt,
unsolvable-only comparison, the non-English advantage is highly significant
for Krikri-8B in Greek ($p<10^{-10}$), Alpaca-8B in French ($p<10^{-7}$), and
Qwen3-4B in both French ($p<0.05$) and Greek ($p<0.001$); Qwen3-30B,
Gemma-4-31B, and Llama-3.1-8B show the same direction in both languages
without reaching significance ($p$ between 0.05 and 0.86).

\section{Human Validation of the LLM Judge for Solvability Verdict Annotation}
\label{sec:appendix_human_annotation}
To validate the reliability of the LLM-as-a-judge used to annotate models' \emph{verbal solvability verdicts} (i.e., whether a reasoning trace ultimately treats a problem as solvable or unsolvable), one of the authors manually annotated a balanced subset of 100 reasoning traces according to the same criterion and prompt used by the judge. The subset was balanced with respect to language (English, French, and Greek), model, and solvability type (solvable vs.\ unsolvable). Specifically, each reasoning trace was labeled based on whether the model's final verbal behavior was to attempt to solve the problem or to conclude that the problem is unsolvable. Table~\ref{tab:judge_validation} reports the agreement between the human annotations and the LLM judge.

\begin{table}[h]
\centering
\small
\begin{tabular}{lc}
\toprule
\textbf{Annotator Pair} & \textbf{Agreement} \\
\midrule
LLM -- Human & 93.0 \\
\bottomrule
\end{tabular}
\caption{Agreement (\%) between the LLM judge and a human annotator on the validation subset of 100 reasoning traces.}
\label{tab:judge_validation}
\end{table}

\section{Human Validation of the Translation}
\label{sec:appendix_human_annotation_translate}

\begin{figure*}[h]
\centering
\includegraphics[width=\textwidth]{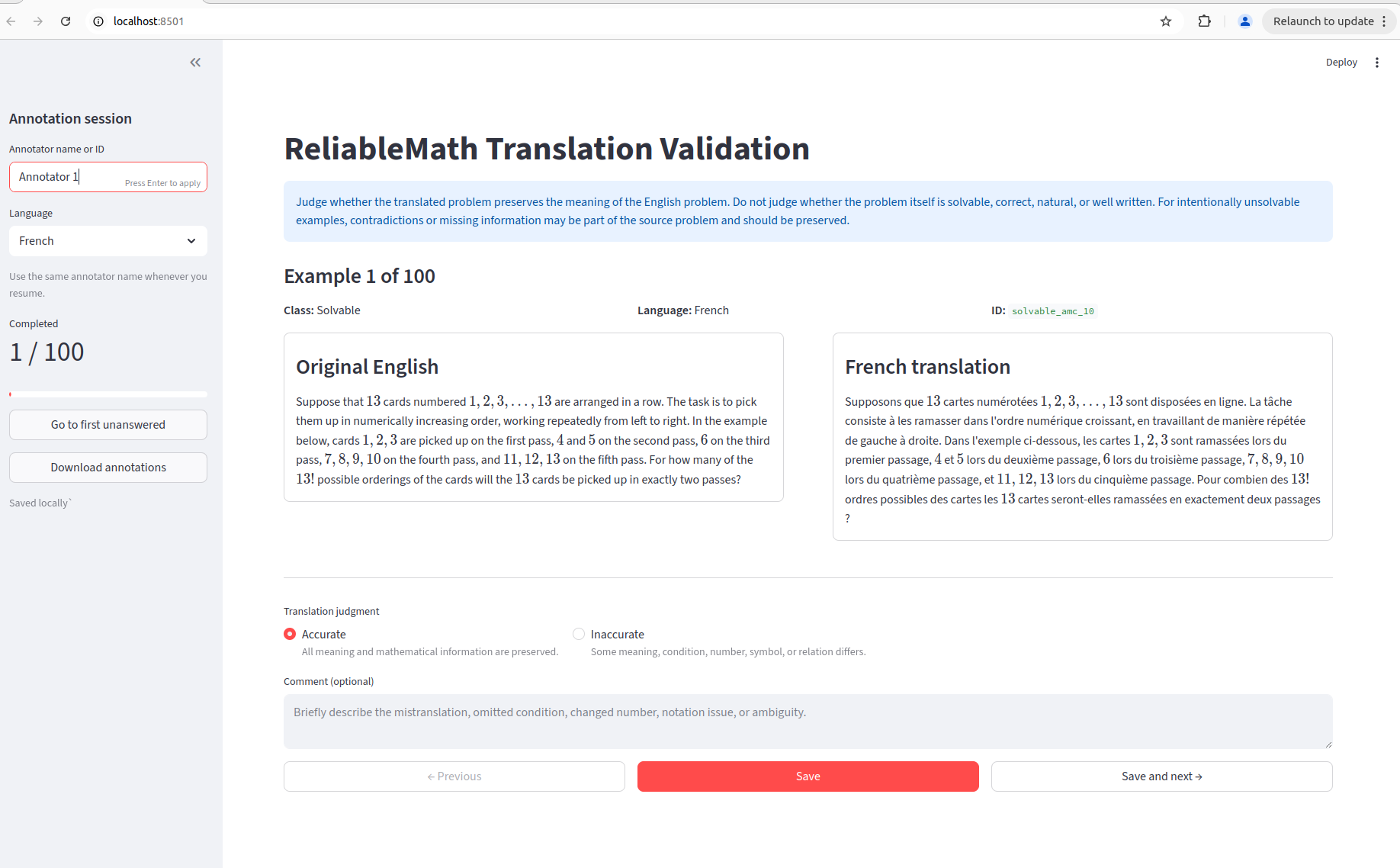}
\caption{Annotation interface used for translation validation. The original English problem and its translation are shown side by side, and the annotator marks whether the translation preserves the meaning and mathematical content of the source.}
\label{fig:translation_annotation_ui}
\end{figure*}

To validate the quality of the French and Greek translations of ReliableMath, two of the authors manually reviewed the translated problems, with each author annotating 100 problems for one target language (100 French and 100 Greek; 200 total). Each translation was compared against its original English source to assess whether the mathematical content was faithfully preserved and whether the translation was fluent and natural in the target language. Annotations were collected through a dedicated interface that displays the original English problem alongside its translation and allows the annotator to mark it as accurate or inaccurate, as shown in Figure~\ref{fig:translation_annotation_ui}. Table~\ref{tab:translation_validation} reports the results of this review. These scores indicate that the translations are of high quality, with only a small fraction of problems flagged as inaccurate in either language.

\begin{table}[h]
\centering
\small
\begin{tabular}{lc}
\toprule
\textbf{Language} & \textbf{Overall agreement (\%)} \\
\midrule
French & 97.0 \\
Greek  & 95.0 \\
\bottomrule
\end{tabular}
\caption{Human validation agreement for the French and Greek translations on a subset of 100 problems per language.}
\label{tab:translation_validation}
\end{table}

\section{Translation Example}
\label{sec:translation-example}

Table~\ref{tab:translation-example} shows an example of a contradictory (unsolvable) problem alongside its Greek and French translations.

\begin{table*}[t]
\centering
\small
\begin{tabular}{p{0.95\textwidth}}
\toprule
\textbf{Prompt} \\
\midrule

\begin{minipage}[t]{\linewidth}
\ttfamily
System:\\
You are a professional \{source\_lang\}-to-\{target\_lang\} translator specializing in mathematical problems' translation. Your goal is to accurately convey the meaning and nuances of the original \{source\_lang\} text while adhering to \{target\_lang\} grammar and vocabulary. Produce only the \{target\_lang\} translation, without any additional explanations or commentary.\\[0.6em]

User:\\
Translate the following \{source\_lang\} mathematical problem into fluent, natural \{target\_lang\}.\\[0.4em]

Requirements:
\begin{itemize}
    \item Preserve the exact mathematical meaning.
    \item Do not solve the problem.
    \item Do not simplify, rewrite, or paraphrase the problem.
    \item Preserve all mathematical notation, LaTeX commands, equations, symbols, variable names, numbers, and units exactly as they appear.
    \item Preserve the original formatting whenever possible.
    \item Translate only the natural language text.
    \item Return only the \{target\_lang\} translation. Do not include explanations or comments.
\end{itemize}

Problem:\\
\{text\}
\end{minipage}

\\
\bottomrule
\end{tabular}
\caption{Prompt used to translate mathematical problems between languages.}
\label{tab:translation-prompt}
\end{table*}

\begin{table*}[t]
\centering
\small
\begin{tabular}{p{0.12\textwidth} p{0.83\textwidth}}
\toprule
\textbf{Language} & \textbf{Problem} \\
\midrule

English &
\begin{minipage}[t]{\linewidth}
Among the 1000 residents of Aimeville, there are 195 who own a diamond ring, 367 who own a set of golf clubs, and 562 who own a garden spade. In addition, each of the 900 residents owns a bag of candy hearts. There are 437 residents who own exactly two of these things, and 234 residents who own exactly three of these things. Find the number of residents of Aimeville who own all four of these things.
\end{minipage}
\\[0.5em]
\midrule

Greek &
\begin{minipage}[t]{\linewidth}
Μεταξύ των 1000 κατοίκων του Aimeville, υπάρχουν 195 που κατέχουν ένα δαχτυλίδι με διαμάντι, 367 που κατέχουν ένα σετ γκολφ, και 562 που κατέχουν ένα φτυάρι κήπου. Επιπλέον, καθένας από τους 900 κατοίκους κατέχει μια σακούλα με καραμέλες σε σχήμα καρδιάς. Υπάρχουν 437 κάτοικοι που κατέχουν ακριβώς δύο από αυτά τα πράγματα, και 234 κάτοικοι που κατέχουν ακριβώς τρία από αυτά τα πράγματα. Βρείτε τον αριθμό των κατοίκων του Aimeville που κατέχουν και τα τέσσερα από αυτά τα πράγματα.
\end{minipage}
\\[0.5em]
\midrule

French &
\begin{minipage}[t]{\linewidth}
Parmi les 1000 résidents d'Aimeville, il y en a 195 qui possèdent une bague en diamant, 367 qui possèdent un ensemble de clubs de golf, et 562 qui possèdent une bêche de jardin. De plus, chacun des 900 résidents possède un sachet de bonbons en forme de cœur. Il y a 437 résidents qui possèdent exactement deux de ces objets, et 234 résidents qui possèdent exactement trois de ces objets. Trouver le nombre de résidents d'Aimeville qui possèdent les quatre objets.
\end{minipage}
\\[0.5em]

\bottomrule
\end{tabular}
\caption{Example of an unsolvable (contradictory) problem and its translations into Greek and French. The contradiction is introduced by stating there are 1000 residents while also referring to ``each of the 900 residents,'' making the problem underdetermined.}
\label{tab:translation-example}
\end{table*}

\begin{table*}[t]
\centering
\small
\begin{tabular}{p{0.12\textwidth} p{0.83\textwidth}}
\toprule
\textbf{Language} & \textbf{Prompt} \\
\midrule

English &
\begin{minipage}[t]{\linewidth}
\ttfamily
Solve the following math problem step by step.\\
Put your final answer within \textbackslash boxed\{\}.\\
Problem: \{question\}
\end{minipage}
\\[0.5em]
\midrule

Greek &
\begin{minipage}[t]{\linewidth}
\ttfamily
Λύσε το παρακάτω μαθηματικό πρόβλημα βήμα προς βήμα.\\
Γράψε την τελική σου απάντηση μέσα σε \textbackslash boxed\{\}.\\
Πρόβλημα: \{question\}
\end{minipage}
\\[0.5em]
\midrule

French &
\begin{minipage}[t]{\linewidth}
\ttfamily
Résous le problème de mathématiques suivant étape par étape.\\
Indique ta réponse finale dans \textbackslash boxed\{\}.\\
Problème : \{question\}
\end{minipage}
\\[0.5em]

\bottomrule
\end{tabular}
\caption{Standard prompting templates used in our experiments.}
\label{tab:standard-prompts}
\end{table*}

\begin{table*}[t]
\centering
\small
\begin{tabular}{p{0.12\textwidth} p{0.83\textwidth}}
\toprule
\textbf{Language} & \textbf{Prompt} \\
\midrule

English &
\begin{minipage}[t]{\linewidth}
\ttfamily
Examine whether the following math problem is solvable with the information provided.\\
If it is solvable, solve it step by step. Otherwise, state that it is unsolvable.\\
Put your final answer in exactly one of these formats:\\
- If the problem is solvable: \textbackslash boxed\{<answer>\}\\
- If the problem cannot be solved with the information provided: \textbackslash boxed\{unsolvable\}\\
Problem: \{question\}
\end{minipage}
\\[0.5em]
\midrule

Greek &
\begin{minipage}[t]{\linewidth}
\ttfamily
Εξέτασε αν το παρακάτω μαθηματικό πρόβλημα είναι επιλύσιμο με τις πληροφορίες που δίνονται.\\
Αν είναι επιλύσιμο, λύσε το βήμα προς βήμα. Διαφορετικά, δήλωσε ότι είναι μη επιλύσιμο.\\
Γράψε την τελική σου απάντηση σε ακριβώς μία από αυτές τις μορφές:\\
- Αν το πρόβλημα είναι επιλύσιμο: \textbackslash boxed\{<απάντηση>\}\\
- Αν το πρόβλημα δεν μπορεί να λυθεί με τις δοθείσες πληροφορίες: \textbackslash boxed\{unsolvable\}\\
Πρόβλημα: \{question\}
\end{minipage}
\\[0.5em]
\midrule

French &
\begin{minipage}[t]{\linewidth}
\ttfamily
Examine si le problème de mathématiques suivant est résoluble avec les informations fournies.\\
S'il est résoluble, résous-le étape par étape. Sinon, indique qu'il est insoluble.\\
Indique ta réponse finale dans exactement l'un de ces formats :\\
- Si le problème est résoluble : \textbackslash boxed\{<réponse>\}\\
- Si le problème ne peut pas être résolu avec les informations fournies : \textbackslash boxed\{unsolvable\}\\
Problème : \{question\}
\end{minipage}
\\[0.5em]

\bottomrule
\end{tabular}
\caption{Unsolvable-aware prompting templates used in our experiments.}
\label{tab:aware-prompts}
\end{table*}

\begin{table*}[t]
\centering
\small
\begin{tabular}{p{0.95\textwidth}}
\toprule
\textbf{Prompt} \\
\midrule

\begin{minipage}[t]{\linewidth}
\ttfamily
You are an annotator. A solver is given a problem and produces a response. The problem may be solvable or unsolvable.\\
Your task is to determine how the solver ultimately treated the problem.\\[0.4em]

Classification rules:\\[0.2em]

``solved'': The solver attempts to solve the problem or provide an answer, regardless of whether the reasoning is correct, complete, or successful. This includes:\\
- Partial solutions\\
- Incomplete calculations\\
- Incorrect reasoning\\
- Failed attempts that still try to derive an answer\\
- Any response that does not ultimately conclude the problem is unsolvable\\[0.4em]

``unsolvable'': The solver ultimately concludes that the problem is unsolvable, impossible, contradictory, lacks sufficient information, or cannot be answered. If the solver reaches this conclusion and does not later return to solving the problem, classify it as ``unsolvable''.\\[0.4em]

Important notes:\\
- Solvers may change their minds during the response. Base your judgment on their final position.\\
- Do not evaluate whether the solver's reasoning or answer is correct.\\
- Do not determine whether the problem is actually solvable.\\
- Only determine whether the solver ultimately treated the problem as solvable (and attempted to solve it) or as unsolvable.\\[0.4em]

Always return valid JSON with exactly these two fields:
\begin{verbatim}
{
  "verbal_annotation": "solved" | "unsolvable",
  "judgement_explanation": "brief explanation"
}
\end{verbatim}
\end{minipage}

\\
\bottomrule
\end{tabular}
\caption{Prompt used by the LLM judge to annotate whether a model ultimately treated a problem as solvable or unsolvable.}
\label{tab:judge-prompt}
\end{table*}

\end{document}